# Automating the Compilation of Potential Core-Outcomes for Clinical Trials

Shwetha Bharadwaj*

sbharadwaj36@gatech.edu

*Georgia Institute of Technology [a]*

Melanie Laffin

laffin_melanie@bah.com

*Booz Allen Hamilton[b], Georgia Institute of Technology*

[a] *North Ave NW, Atlanta GA*
[b] *901 15th St NW, Washington DC*

## Abstract

Due to increased access to clinical trial outcomes and analysis, researchers and scientists are able to iterate or improve upon relevant approaches more effectively. However, the metrics and related results of clinical trials typically do not follow any standardization in their reports, making it more difficult for researchers to parse the results of different trials. The objective of this paper is to describe an automated method utilizing natural language processing in order to describe the probable core outcomes of clinical trials, in order to alleviate the issues around disparate clinical trial outcomes. As the nature of this process is domain specific, BioBERT was employed in order to conduct a multi-class entity normalization task. In addition to BioBERT, an unsupervised feature-based approach making use of only the encoder output embedding representations for the outcomes and labels was utilized. Finally, cosine similarity was calculated across the vectors to obtain the semantic similarity. This method was able to both harness the domain-specific context of each of the tokens from the learned embeddings of the BioBERT model as well as a more stable metric of sentence similarity. Some common outcomes identified using the Jaccard similarity in each of the classifications were compiled, and while some are untenable, a pipeline for which this automation process could be conducted was established.

**Introduction**

Medical and regulatory communities have made it possible for the increase in the proliferation and access to analysis involving clinical studies by way of prepublications and embargos. This in turn helps researchers and other experts iterate or improve on relevant approaches and in a more general sense, fosters a sense of collaboration in a time of mandated separation.

Although this torrent of information via relaxed protocol surrounding publication is vital for progress, it unfortunately also has some drawbacks including lack of standardization of the clinical trial results and metrics. This discrepancy between disparate clinical trial results is not a new problem. Dodd et. al recognized such an issue when reporting on core outcome datasets after analyzing the core outcome measures in the effectiveness trials database. To ensure discovery of disparity between core outcome sets within the database, a common taxonomy was devised that both encompassed the granularity and comprehensiveness of clinical outcomes across studies.

The purpose of this paper is the creation of an automation framework to reduce the amount of manual oversight required by domain experts to create these sets of outcomes. Given a database of outcomes from specific clinical studies and a comprehensive taxonomy (similar to the ones produced by Dodd et. al), outcomes are grouped into their respective classifications, with recurrent ones required for the standardized set. Then, the oversight by professionals might be limited to this parsed set rather than the entire database of outcomes reported by each study. The utility of these experiments can be seen as a proof of concept that the outcomes can be categorized using an unsupervised machine learning approach. This can be used as a possible way of filtering the most pertinent outcomes per classification, creating a potential standardized metric. Therefore, it is possible less duplication of efforts may occur as clinical trial researchers could obtain more meaningful results when searching for outcomes of clinical trials. In addition, the unsupervised approach allows for a more data-driven first pass with the potentiality for supervisory scrutinization by experts for the more contentious labels.

## Methods & Results

This article tests two main taxonomies in order to classify outcomes taken from the AACT database, related to "covid-19". Specifically, outcomes regarding the query 'covid-19' were scraped as free text and classified by each category within 2 main taxonomies:
- The Smith et. al 15-class taxonomy for outcomes classified in the Cochrane Reviews.
- A 5-core area version.

The task of classifying these unlabeled outcomes into a pre-specified ontology can be viewed as an entity normalization task as described by Ji et. al. This paper builds on the work of Ji et. al, notably differences including:
- each outcome is mapped to exactly one classification (i.e., a multi-classification) without a possibility of an 'unlinkable entity mention'.
- The knowledge base in this paper is fixed (based solely on the proposed taxonomies by domain-experts).
- a 'one-vs-rest' multi-class classification approach, as opposed to computing the SoftMax score over all possible concepts.

The clinical trial design outcomes were scraped from the AACT database - "a publicly available relational database that contains all information (protocol and result data elements) about every study registered in ClinicalTrials.gov."

### *Use of BioBERT*

Natural language processing (NLP) in the biomedical domain typically finds issues in the specificity in terms of domain vocabulary. Prior solutions have incorporated pre-trained embedding vectors for biomedical-specific corpora for their respective tasks, including concept embeddings (CUIs) and Medical Subject Headings (MESH) word vectors. Although standardized in terms of usage and used in a swath of prior work with regards to natural language processing in the biomedical subdiscipline, its rigidity in terms of its pre-defined vocabulary prevents tasks with less researched and studied terms. In other words, the corpora might not be updated with newer diseases and preliminary findings.

To overcome this hurdle, BioBERT, a bi-directional transformer model utilizing a word-piece tokenization schema and therefore a contextual representation is used. While it utilized weights from BERT which was pre-trained on general domain corpora including Wikipedia and Bookscorpus, it was further pre-trained on PubMed abstracts and full-text articles for 23 days on 8 NVIDIA V100 GPUs attesting to its domain-specific relevance.

Because the nature of this process is domain specific, BioBERT was employed in order to conduct multi-class entity normalization tasks, or how to best classify each of the outcomes into a category within an ontology given their contextual embeddings. The specific BioBERT model used was the BioBERT-Large v1.1 (+ PubMed 1M) - based on BERT-large-Cased (custom 30k vocabulary). The Estimated size of each corpus is 4.5 billion words for the PubMed abstracts, and 13.5 billion words for PubMed Central Full Texts.

*Utilizing Unsupervised Learning*

The entity normalization task was first carried out using the pre-existing weights from the BioBERT-Large v1.1 model. As such, it was solely a downstream task that consisted of sentence classification without any domain-specific training. For 100 COVID-19 clinical trial study outcomes scraped from the AACT database, the following probability distribution from the SoftMax layer for the classification task amongst the Smith taxonomy were acquired.

|  | Counts |
|---|---|
| **operative** | 94 |
| **adverse events effects** | 4 |
| **quality life** | 2 |

*Figure 1*

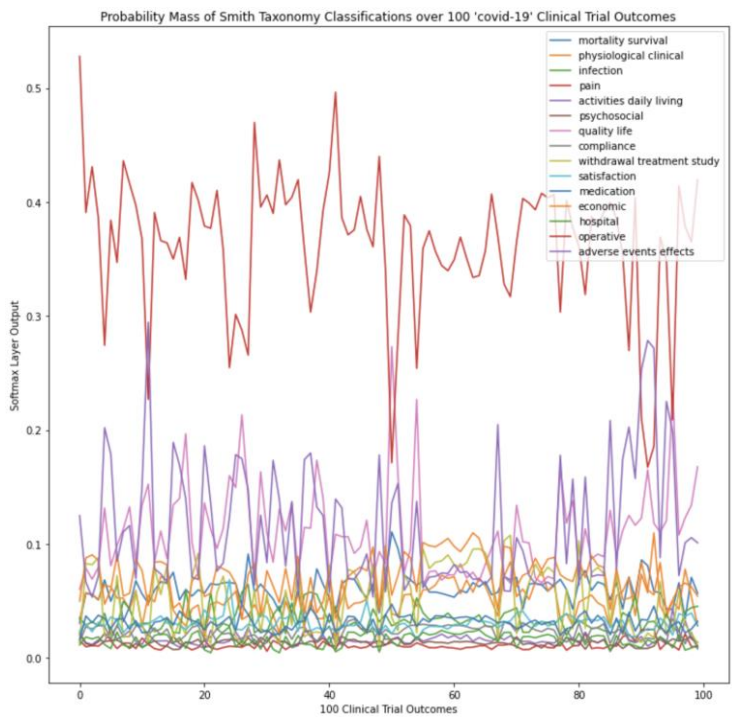

*Figure 2*

Obviously, the results from this distribution are extremely problematic as the BioBERT architecture is assigning each outcome to the 'operative' label when clearly, this should not be the case. For example, the outcome, 'mortality rate' should be assigned the label 'mortality survival'. We can visualize where the BioBERT model is placing its attention by obtaining the last transformer block and averaging across the multi-headed attention layer over the attention heads for each of the tokens' attention weights inputting into the [CLS] token.

Predicted label: operative
Probabilities: [0.0648,0.0346,0.0427,0.0147,0.0200,0.0177,0.1406,0.0223,0.0218,0.0253,0.0208,0.0493,0.0076,0.3927,0.1250]
[CLS] mortality rate [SEP]

Although both 'mortality' and 'rate' had a high attention weight score (as characterized by the darkness of the hue), the label 'operative' was still predicted. The second highest was 'adverse events effects', and only then was the label 'mortality survival' predicted.

For the 'core areas outcomes' taxonomy, the following counts and respective probability mass distribution was acquired:

| | Counts |
|---|---|
| death | 62 |
| physiological clinical | 30 |
| life impact | 8 |

*Figure 3*

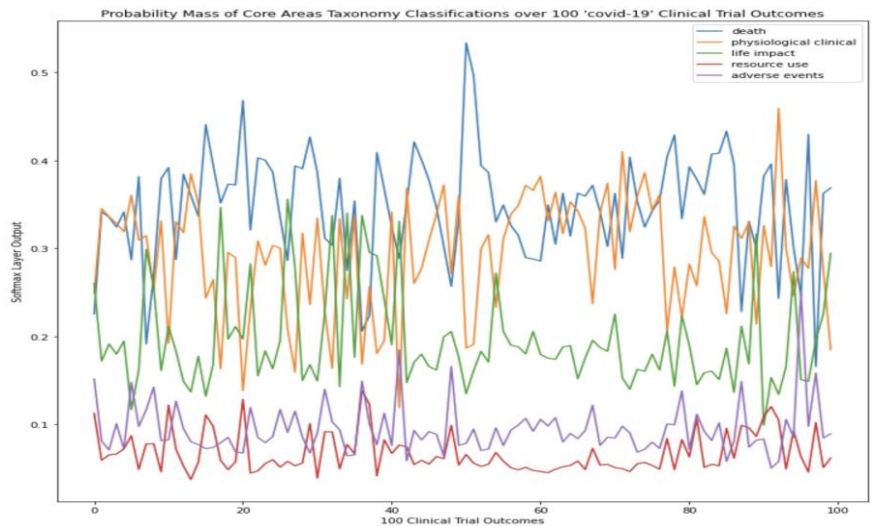

*Figure 4*

## Unsupervised Text Classification After Pre-Training on 10,000 'covid-19' Abstracts

One solution to this problem was to pretrain the model on a different set of corpora in order to incorporate learned subword tokens within the classification task. A context-oriented approach by scraping 10,000 abstracts from PubMed central relating to the query 'covid-19'. These articles were then each separated into sentences to be fed into the model for their respective pre-training tasks: multi-masked language modeling and next sentence prediction.

Utilizing pre-trained model with the same prediction hyperparameter setup and acquired the following results for the Smith Taxonomy:

|  | Counts |
|---|---|
| withdrawal treatment study | 26 |
| infection | 25 |
| satisfaction | 17 |
| pain | 14 |
| activities daily living | 12 |
| psychosocial | 2 |
| operative | 2 |
| mortality survival | 1 |
| economic | 1 |

*Figure 5*

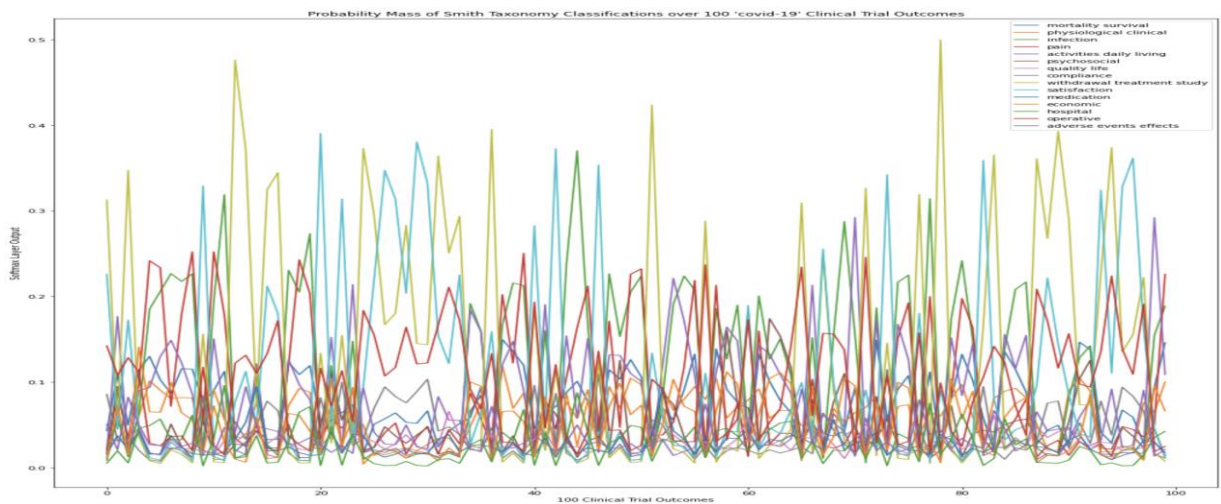

*Figure 6*

As seen above, the distribution of assigned outcomes is much more widely varied than for the un-pretrained model, potentially because the embeddings learned for the subword tokens were attended to more in each of the attention head functions. However, looking again at the visualization, the separator token again was highly attended to.

The same paradigm emerged for the core area taxonomy as well: Although the outcomes were assigned from the full ontology of labels, the separator tokens were widely attended to and results of the classification were either extremely counterintuitive or contradictory.

|  | Counts |
|---|---|
| life impact | 28 |
| death | 28 |
| resource use | 18 |
| adverse events | 13 |
| physiological clinical | 13 |

*Figure 7*

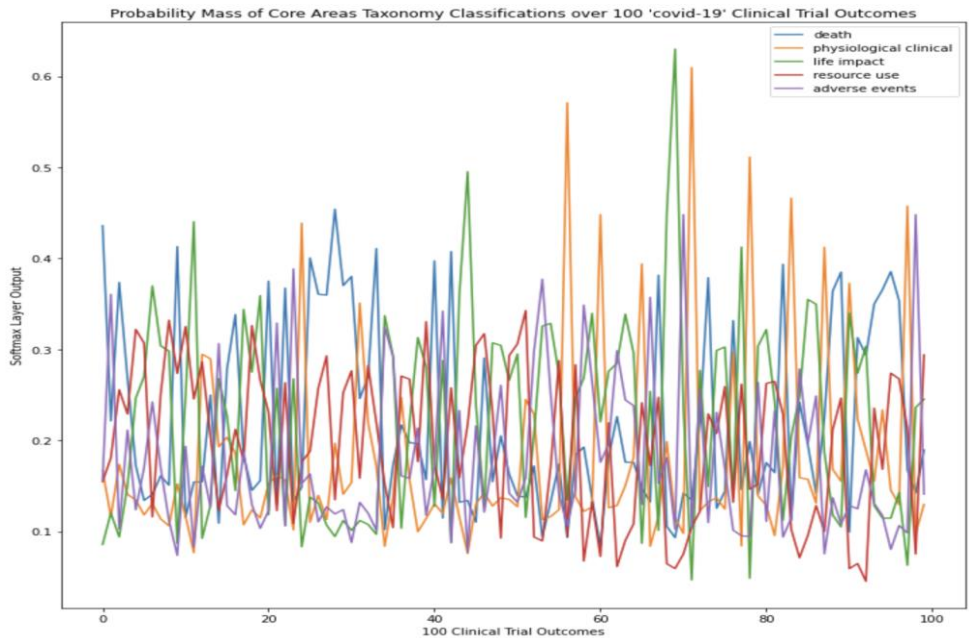

*Figure 8*

*Utility of the Un-Trained BioBERT model for downstream tasks*

The main problem as seen from large attention placed on the separator tokens can be boiled down to the model not able to find any context within the outcome on which to classify one of the pre-ordained labels. As such, it is pertinent to either treat this task as an unsupervised categorization or clustering. One can leverage the encoder outputs from BioBERT as embeddings and utilize a 'Feature Based Approach' as described by Devlin et. al

In an attempt to mitigate this issue, cosine similarity was utilized to categorize the outcomes into respective classifications from the two taxonomies. No task-specific architecture supplemented the output embeddings. Pretrained BioBERT models were used on the additional PMC abstracts with a hidden layer (and therefore embedding dimension) of 1024. The median of the embeddings of the output tokens was used as the final embedding for each outcome. In short, we employed the encoder embeddings output from the model for each outcome as our feature vectors and then assigned the highest cosine similarity score between the outcomes and labels as the classification.

Below are the counts of classifications for 100 random outcomes from the AACT database relating to COVID-19 studies classified under the Smith et. al Taxonomy:

|  | Counts |
| --- | --- |
| withdrawal treatment study | 28 |
| activities daily living | 27 |
| adverse events effects | 16 |
| quality life | 14 |
| psychosocial | 7 |
| satisfaction | 4 |
| physiological clinical | 3 |
| mortality survival | 1 |

*Figure 9*

Only 8 out of the 15 classifications were utilized in this smaller scale classification, however, this is similar to the spread of classifications obtained using the untrained weights from the SoftMax prediction. Clustering by cosine similarity yielded more intuitive results regarding the same outcomes. For example, the outcome 'mortality rate' yielded the outcome of 'mortality survival' as its top-ranking classification rather than that of 'pain' as obtained by the SoftMax classifier on the un-fine-tuned weights. The outcomes 'adverse events absolute number' and 'adverse events percentage' also were mapped to the more rational classification 'adverse events effects' rather than 'pain' or 'infection'.

| Outcome | mortality rate |
|---|---|
| Rank 1 Classification | (mortality survival, 0.08427025809403207) |
| Rank 2 Classification | (withdrawal treatment study, 0.0820156684211132) |
| Rank 3 Classification | (quality life, 0.07345215514029457) |

*Figure 10*

|  | Outcome | Rank 1 Classification | Rank 2 Classification | Rank 3 Classification |
|---|---|---|---|---|
| 18 | adverse events absolute number | (adverse events effects, 0.10470478514771547) | (activities daily living, 0.07406169796653218) | (physiological clinical, 0.07151622603744455) |
| 19 | adverse events percentage | (adverse events effects, 0.11745326427881052) | (physiological clinical, 0.07020435735201251) | (withdrawal treatment study, 0.06946365723241006) |

*Figure 11*

The outcome 'mortality' unfortunately was classified under 'satisfaction', but as we can see by the rankings, 'mortality survival' was a close second. One can surmise pretraining the model on an even wider corpora might have yielded a stronger threshold between the two classifications.

| Outcome | mortality |
|---|---|
| Rank 1 Classification | (satisfaction, 0.07983480404141685) |
| Rank 2 Classification | (mortality survival, 0.07842027449192909) |
| Rank 3 Classification | (operative, 0.07573294216734362) |

*Figure 12*

Shown below is the Euclidean distance between the cluster centers (mean embeddings) and the embedding vectors for the Smith taxonomy. We can infer that the proximity between categories align with their semantic similarity (expressed in their cosine similarity) to the category:

| Distance Between Cluster Mean and Classifcation Embedding | |
|---|---:|
| activities daily living | 10.556162 |
| adverse events effects | 8.446987 |
| mortality survival | 13.351438 |
| physiological clinical | 10.939546 |
| psychosocial | 11.750631 |
| quality life | 9.547534 |
| satisfaction | 5.847628 |
| withdrawal treatment study | 9.093685 |

*Figure 13*

Below is the 3D t-SNE representation, with distances from the classification vectors to their respective categorized outcome highlighted:

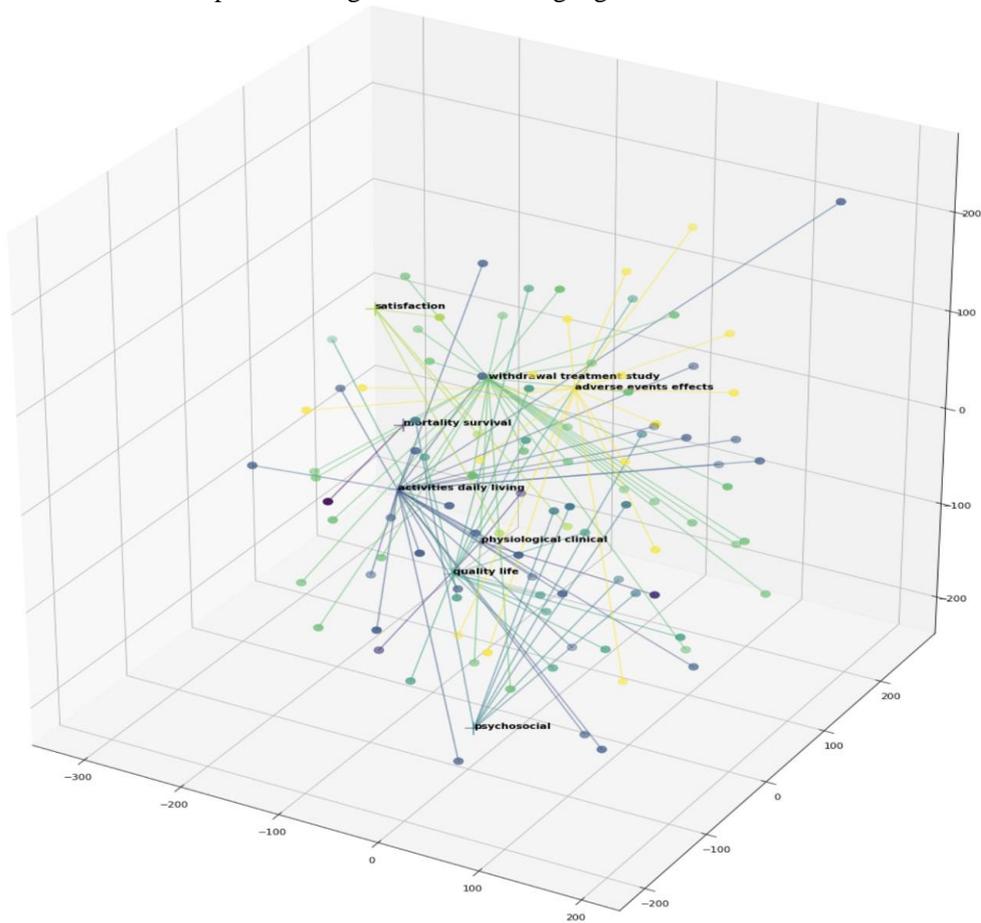

*Figure 14*

Below are the counts for the core area taxonomy classification for the same 100 outcomes:

| | Counts |
|---|---|
| life impact | 32 |
| resource use | 29 |
| physiological clinical | 22 |
| adverse events | 13 |
| death | 4 |

*Figure 15*

As an aside, the core area outcomes as a classification system are seemingly opaque, especially with regards to the breakdown of outcomes within the COMET database, a broad range of COS can be classified within each of the core area outcomes due to its generality.

Although again the cosine similarity metric provided a much more intuitive classification schema for the problematic outcomes identified using the previous approach,

| | Distance Between Cluster Mean and Classifcation Embedding |
|---|---|
| adverse events | 9.910472 |
| death | 5.694263 |
| life impact | 9.894615 |
| physiological clinical | 10.824276 |
| resource use | 9.662728 |

*Figure 16*

| | |
|---|---|
| Outcome | mortality rate |
| Rank 1 Classification | (resource use, 0.21465679828050502) |
| Rank 2 Classification | (life impact, 0.21320261173049476) |
| Rank 3 Classification | (death, 0.202416361775799) |
| Classification | resource use |

*Figure 17*

| | |
|---|---|
| **Outcome** | mortality |
| **Rank 1 Classification** | (death, 0.265424776894284) |
| **Rank 2 Classification** | (physiological clinical, 0.19984455731845988) |
| **Rank 3 Classification** | (life impact, 0.18518266099522376) |
| **Classification** | death |

*Figure 18*

| | |
|---|---|
| **Outcome** | ratio |
| **Rank 1 Classification** | (death, 0.26257773852696814) |
| **Rank 2 Classification** | (physiological clinical, 0.21767549116807208) |
| **Rank 3 Classification** | (life impact, 0.1769397994870675) |
| **Classification** | death |

*Figure 19*

A future goal may be how best to assess a certain threshold by which further expertise is needed to apply a more accurate classification.

Figure 20 shows the t-SNE visualization for the 5-point core-area outcome classification. Because of the smaller cardinality of the classification set, the clustering structure is much more compact, suggesting a better 'grouping'. However, the set's lack of comprehensiveness is evidenced by the outstretched outliers.

Finally, the above computational pipeline was executed for a larger outcome set (24,000 scraped outcomes from the AACT database) and some we identified some common outcomes discovered using the Jaccard similarity index. Counts of classifications for each of the taxonomies are shown in Figure 21.

The pairwise Jaccard similarity for outcomes labelled under each classification was calculated, and the ones with the least differences are shown below. Note that these are not the MOST frequently occurring outcomes and rather a random subset of frequently occurring outcomes as ranked by similarity with one another.

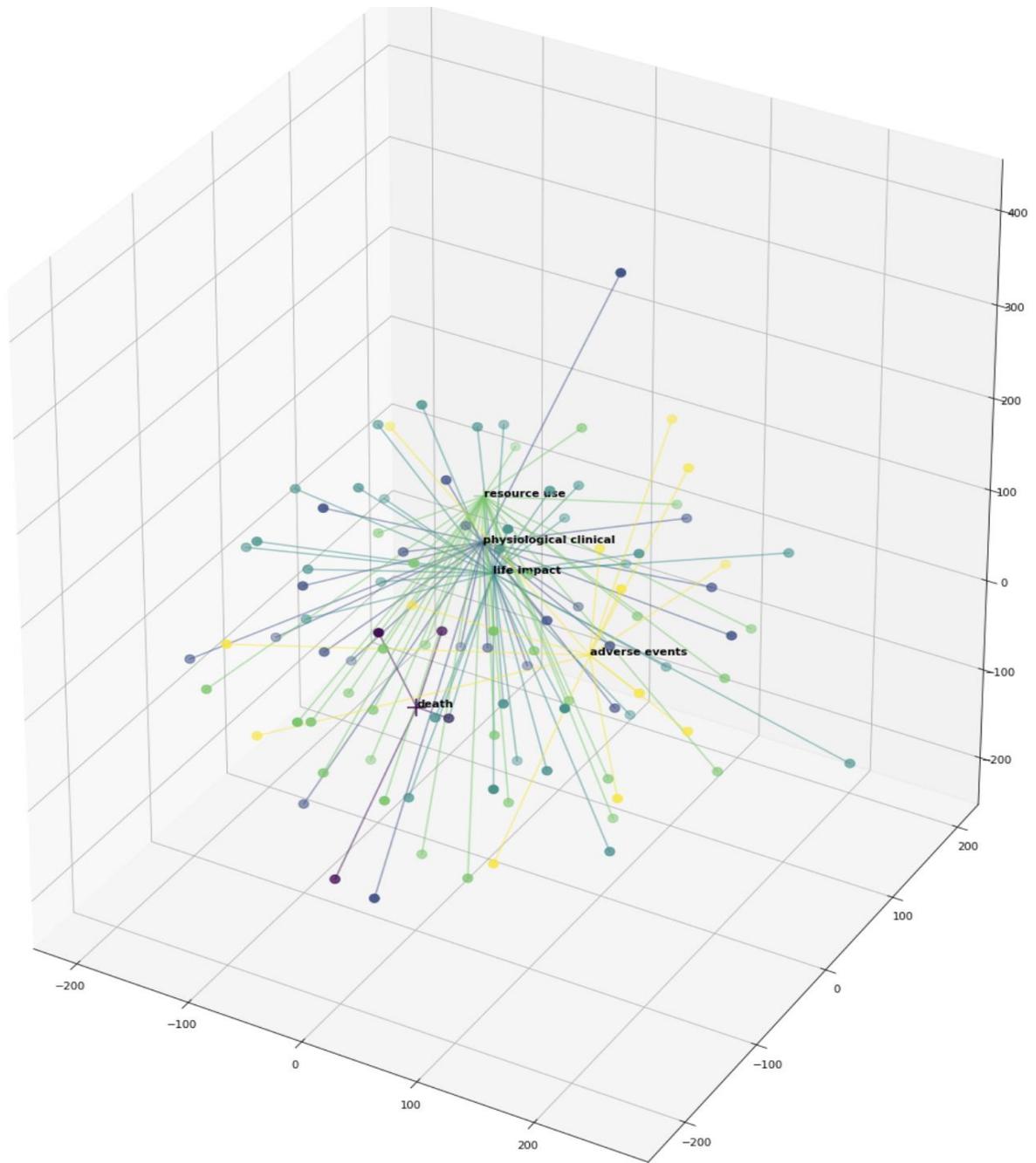

*Figure 20*

|  | Counts |
|---|---|
| withdrawal treatment study | 10100 |
| activities daily living | 4775 |
| adverse events effects | 3196 |
| quality life | 3139 |
| satisfaction | 600 |
| psychosocial | 552 |
| physiological clinical | 538 |
| mortality survival | 455 |
| compliance | 174 |
| operative | 103 |
| pain | 102 |
| economic | 90 |
| hospital | 34 |
| infection | 22 |
| medication | 20 |

|  | Counts |
|---|---|
| life impact | 10461 |
| resource use | 6806 |
| physiological clinical | 3087 |
| adverse events | 2045 |
| death | 1501 |

*Figure 21*

## Discussion

As per the examples shown and others not presented in the figures, the separator token was largely attended to. As per Clark et. al, this can be seen as the attention head function not being applicable and consequently attention to the [SEP] token as a no-op. In other words, because the label itself did not find any token 'worthy' of attention (high similarity), it gave a predominant weighting to the separator token.

One line of reasoning for the poor prediction quality was the wordpiece tokenization schema implemented. While trained on a large corpora within the biomedical domain, the vocabulary itself of BioBERT is a little outdated especially with respect to jargon relating to 'covid-19' queries for publications. As shown below, 'covid' is not a token within the BioBERT vocabulary and is regressed to subwords. In addition, several other seemingly generic biomedical terms are also fragmented, e.g., 'brainstem' and 'sedation'. While only incorporation of these terms into the vocabulary would solve the fragmentation issue, one can pre-train these wordpieces on a variety of corpora to gain contextual embeddings for those specific tokens. Although there might be conflicting interpretations due to several words containing these subtokens, should specific wordpiece tokens be used often enough in correlations with specific words, then those contextual embeddings might be represented on a pretrained version.

As demonstrated by Williamson et. al, the necessity of standardization of core outcome datasets for clinical trials is evidenced by both omission of potentially substandard results as well as duplication of results from unawareness of previously conducted studies. Should a standard of outcomes be required for subdisciplines of scientific trials, a comprehensive and unbiased view of the result set would be reported instead of selective ones revealing only the 'better' outcomes. In addition, by establishing a set database of outcomes, researchers can acquire a better understanding of existing experiments and leverage this information for iterative studies, thereby fostering improvement rather than redundancy.

Embeddings of the BioBERT model as well as a more stable metric of sentence similarity, as the attention schema seemed to be lacking. The latter was evidenced by our results in which many of the previously problematic outcomes being classified as erroneous labels were now classified with more reasonable ones.
Some common outcomes identified using the Jaccard similarity in each of the classifications were compiled, and while some are still untenable, a pipeline for which this automation process could be conducted was established with this last attempt.

Some future goals might be to try to better understand the attention mechanism being employed, more specifically as to why the '[SEP]' token dominated the distribution of weight scores. This might be explained in part by a lack of a 'training set' in which the SoftMax layer might have learned weights after a training procedure instead of the standard weight initialization. Another goal would be to construct a metric by which the accuracy of these classifications might be held against. Some thoughts include the variance of the top three results in the figures: Outcomes with close classification scores might be scrutinized and thus awarded a lower accuracy. The distance between the embeddings of the classifications and outcomes within each cluster might also be used to see which outcomes might be qualified as outliers.

# Acknowledgements

The authors would like to acknowledge the guidance and help of Ken Wilkins, Susan Dodd, and Michael Abaho.

*Ethical Statement*

This material is the authors' own original work, which has not been previously published elsewhere. The paper is not currently being considered for publication elsewhere. The paper reflects the authors' own research and analysis in a truthful and complete manner.

*Statement of Support and Competing Interests*

This material was not supported by any funding or grants, nor are there any competing interests in the material of this paper.